\documentclass[12pt]{article}

\usepackage{amsmath,amssymb,amsfonts,amsthm}
\usepackage{mathrsfs}
\usepackage{booktabs}
\usepackage{subfig}
\usepackage{graphicx}
\usepackage{float}
\usepackage{tabularx}
\usepackage{enumerate}
\usepackage{enumitem}
\usepackage{url}
\usepackage{natbib}
\usepackage{hyperref}
\newtheorem{theorem}{Theorem}[section]
\usepackage{setspace}

\begin{document}

  \title{\bf Utility-Aware Multimodal Contrastive Learning for Product Image Generation}

  \author{Xiaohang Feng, 
  Yiling Xie\\
  City University of Hong Kong}
  \date{}

\maketitle

\begin{abstract}
Product images strongly influence consumer decision-making in online marketplaces. Empowered by multimodal contrastive learning, generative AI can output images that closely align with text prompts. Yet existing generative AI models do not directly optimize marketplace performance. This is a critical gap, since semantic alignment alone does not guarantee that an image will sell. To address this limitation, we propose a \textit{utility-aware multimodal contrastive learning} framework that incorporates consumer demand into a novel Utility-Aware InfoNCE loss. Optimizing this utility-aware objective guides generation toward images that are both semantically coherent and demand-enhancing. This effect arises directly from a shift in the learned image-text representation space toward demand-driven visual cues, which we also validate through the theoretical bound of the proposed objective.
In downstream applications on Amazon and Airbnb, product images generated and edited by our method outperform state-of-the-art models in increasing demand and preserving fidelity, while maintaining text–image consistency. Notably, our utility-aware framework preserves inverse U-shaped demand patterns for attributes such as aesthetics and uniqueness, improving demand-based performance while preserving fidelity and semantic consistency. Human-subject experiments further validate its commercial effectiveness. As generative AI technology continues to evolve, our utility-aware component can be flexibly embedded into emerging generative models to improve direct commercial use. 
\end{abstract}

\setstretch{1.5}
\section{Introduction}

Visual content and product images are economically valuable assets on digital platforms because they shape consumer attention, trust, and purchasing behavior. 
The prior literature demonstrates that the characteristics of the image significantly affect the demand in online marketplaces \citep{zhang2022airbnb,li_xie_2020}. Industry evidence further corroborates this: Airbnb reports that listings with professional photos receive 19\% more bookings \footnote{\url{https://www.airbnb.com/e/pro-photography}}, while Salsify reports that 71\% of shoppers have returned an item due to incorrect product content online \footnote{\url{https://www.salsify.com/resources/report/2025-consumer-research}}. These figures suggest that visual content functions as an economic investment, generating measurable value through increased demand and improved seller profit.

Recent advances in Generative AI, especially text-to-image models, have significantly expanded firms’ ability to create product visuals at scale. A key breakthrough is Contrastive Language-Image Pretraining (CLIP), which bridges text and image modalities. CLIP and its variants have become core components in emerging AI models for image generation \citep{radford2021clip}. Beyond generation, CLIP-based measures, such as the CLIP score, are also widely used to evaluate image–text alignment \citep{hessel2021clipscore}.

The best product image should not be the prettiest one but the one that sells, and these two types of images are almost never the same. The problem is that current AI systems, despite their strong semantic matching capabilities, are optimized to produce the former rather than the latter, which makes them costly and inconvenient for direct business use.
To push these general AI models toward commercial output, users must rely heavily on prompt engineering, repeatedly refining long prompts in a way that increases token costs and reduces efficiency. Yet even after this effort, the results remain noisy and unstable since small wording changes can generate inconsistent or unrealistic images. The models also tend to overemphasize aesthetics, producing over-stylized outputs that deviate from realistic product representation and may weaken consumer trust.
More fundamentally, the AI existing systems are tuned for image–text coherence instead of marketplace performance. This concern is increasingly echoed in industry. A growing view holds that general-purpose generative AI tools fall short in marketing because they are trained to imitate human style rather than to maximize business outcomes, and therefore cannot distinguish between content that is merely eye-catching and content that drives clicks, conversions, or retention.

To address this gap, we study how to generate \textit{utility-aware} product images that drive up demand without extensive prompt engineering. By ``utility-aware", we mean that image generation is guided by both text-image coherence and the demand model that links image features to marketplace performance. We ask three research questions. \textit{First}, how can the demand model be used to construct a utility-aware objective for multimodal contrastive learning? This question studies how demand-driven image features and their estimated coefficients can be integrated into multimodal contrastive learning. \textit{Second}, why does the proposed utility-aware framework work? This question examines the theoretical properties and interpretability of the framework. It clarifies how a demand-based objective changes the learned multimodal representation, why the resulting representation is more aligned with demand-driven image variation, and how this alignment guides generation with greater marketplace relevance. \textit{Third}, can the proposed framework generate utility-aware images across business contexts? This question evaluates whether the method produces images that improve demand-driven performance on e-commerce and rental platforms.

To answer these questions, we proceed in four steps. \textbf{First, method development.} We propose a new  {Utility-Aware InfoNCE loss} that incorporates demand-driven objectives into multimodal contrastive learning. The key idea is to regularize standard image-text similarity score by utility-driven visual attributes. Starting from pretrained CLIP weights, we train the model by optimizing our new objective, Utility-Aware InfoNCE Loss, and refer to the resulting model as \textit{Utility-Aware CLIP}. This step defines the learning objective and the training procedure for producing utility-aware image-text representations. 

\textbf{Second, interpretation.} We explain why the utility-aware framework works. We show how the new loss changes standard contrastive learning. We use theoretical bounds and occlusion-based interpretation to show that Utility-Aware CLIP still preserves image-text consistency, while attending more to demand-driven visual cues.

\textbf{Third, empirical application.} We establish an image generation model based on the Utility-Aware CLIP, which we name the \textit{Utility-Aware Generator}. We then apply the Utility-Aware Generator to both image generation and image editing in Amazon and Airbnb settings. In both contexts, the estimated demand models reveal inverted U-shaped preferences, implying that effective images should be visually appealing and distinctive, but not overly stylized. Guided by these demand patterns, our utility-aware framework consistently outperforms the base model without Utility-Aware CLIP and other baselines, including GPT-Image \cite{chatgpt_image}, Stable Diffusion \citep{podell2023sdxl}, and Flux \citep{flux-2-2025}, across demand-based scores and related performance metrics. Three findings emerge: (1) the utility-aware framework improves both generation and editing outcomes in terms of demand-based performance; (2) it better preserves product identity or listing realism while maintaining semantic consistency; and (3) it avoids the over-polished, over-stylized, or unrealistic outputs that generic image-generation models often produce.

\textbf{Fourth, human validation.} We conduct two human-subject experiments to validate whether the generated images increases demand. In the Amazon setting, participants choose the product image that they would be most likely to purchase. In the Airbnb setting, participants choose the property image that they would be most likely to book. The results validate that Utility-Aware Generator produces images that consumers are more likely to purchase. 

To summarize, this paper proposes a Utility-Aware CLIP framework that incorporates estimated consumer demand into multimodal contrastive learning. Theoretically, it will be shown that our utility-aware framework can be connected to discrete-choice utility maximization. Empirically, across Amazon and Airbnb settings, our method can generate more effective images than existing models by improving realism, demand relevance, and consumer preference outcomes.

These findings provide several managerial implications. For sellers and hosts, the framework provides a scalable way to improve product and property images without relying on repeated prompt experimentation or costly manual design. Instead of asking users to guess the right prompt, the method uses observed marketplace data to guide image generation toward visual attributes that are more likely to improve demand. For example, an Amazon seller can generate additional usage-scenario images for a lamp, blanket, or table while preserving the original product design and avoiding overly polished scenes that make the product look unrealistic. Similarly, an Airbnb host can improve the image of a bedroom or living-room by adding moderate decorative elements, warmer lighting, or more inviting textures, while preserving the layout and credibility of the property. For platforms, the approach provides a systematic tool for content improvement, listing enhancement, and counterfactual testing. Platforms can use our {Utility-Aware Generator} to recommend improved images to sellers, identify listings with low visual quality, or test alternative product presentations before deployment. For example, Amazon could suggest a more demand-aligned secondary image for a product listing, while Airbnb could recommend edits that make a room appear warmer and more distinctive without making it look artificial. The framework can also support sellers with fewer creative or technical resources by reducing the need for professional photography, design expertise, or heavy prompt engineering.

Importantly, our contribution is modular.  
Our proposed  utility-aware contrastive learning framework can be flexibly embedded in emerging AI generative systems to improve their direct commercial use. 
Our proposed demand-guided module can help tailor Generative AI toward images that are not only consistent with text prompt but more importantly, aligned with marketplace performance.

\section{Literature Review}

Our paper is related to four streams of research, namely visual analytics in marketing, Generative AI in marketing, image generation models, and regularized learning. 

\subsection{Visual Analytics in Marketing}

Visual analytics in marketing has evolved from treating images as unstructured content to modeling them as scalable, theoretically grounded inputs to understand consumer behavior and marketplace outcomes. Early work shows that visual data can be used to measure brand meaning itself: \citet{dzyabura2021visual} combines respondent-created collages with machine learning to recover brand associations at scale. Subsequent research links naturally occurring visual content to concrete marketing outcomes. \citet{hartmann2021brand} uses computer vision to classify brand imagery on social media and show that the perspective of the image shapes different forms of engagement. \citet{zhou2021video} extends this logic to video, proposing a computer-vision-based feature framework that predicts online video consumption. \citet{zhang2022airbnb} shows that interpretable visual features extracted from Airbnb listing photos predict demand, while \citet{dzyabura2023returns} demonstrates that prelaunch product images improve return-rate prediction and assortment decisions. \citet{ceylan2024photos} further shows that photos increase the usefulness of the review, particularly when visual and textual information are congruent.

Recent work increasingly develops marketing-specific visual constructs rather than relying only on generic object recognition. \citet{feng2025cvp} develops a machine-learning measure of celebrity visual potential from facial features , and \citet{feng2025uniqueness} constructs and validate a visual uniqueness measure, documenting an inverted U-shaped relationship between uniqueness and Airbnb demand. Relatedly, \citet{zhang2025smile} shows that a host's smile in profile photos increases Airbnb demand, and \citet{zhou2026color} identifies color saturation as a status cue in luxury branding. Taken together, this literature shows that current practice in marketing is moving toward interpretable, domain-specific visual measurement that links images to economically meaningful outcomes such as engagement, demand, returns, and perceived status. Our paper also falls in this stream of work, but we are among the very first to explore Generative AI in visual marketing. 

\subsection{Generative AI in Marketing}

Early studies show that LLMs can approximate consumer preference structures and brand perceptual maps from text data in ways similar to traditional survey methods \citep{gui2023challenge}. More recent research fine-tunes LLMs to generate marketing emails and evaluates their performance in large-scale field experiments \citep{angelopoulos2024causal}. Related work also fine-tunes image generation models on marketing-relevant objective metrics, showing that AI-generated visual content can match or even outperform conventionally produced advertising content in downstream performance outcomes \citep{Heitmann2025Picture}. In addition, LLMs have been integrated into adaptive experimentation frameworks to dynamically optimize content delivery over time \citep{doi:10.1287/mksc.2024.0990}. Our paper focuses on the latest Generative AI models, but we are among the very first to explore Generative AI in visual marketing. 

\subsection{Image Generation Models} 
Recent years have witnessed a major breakthrough in text-to-image generation, led by powerful models such as GPT-Image \citep{chatgpt_image}, Stable Diffusion \citep{podell2023sdxl}, and Flux \citep{flux-2-2025}. These models have demonstrated excellent capabilities in generating high-quality visuals and following complex text prompts. Among existing image generation models, Flux represents the current state-of-the-art open-source image generation model and is chosen as our base model. 

Existing image generation models primarily optimize for visual realism and semantic text-image alignment, but they do not directly account for whether an image improves purchase likelihood, bookings, or sales. Our utility-aware model addresses this gap by learning demand-driven visual preferences from marketplace data and embedding them into the generation process. As a result, it produces images that are not only prompt-consistent and visually plausible, but also aligned with downstream economic outcomes. This shifts image generation from an aesthetics-driven task to a demand-aware tool for marketplace applications.

Similarly, evaluation methods such as CLIP-based similarity metrics improve measurement of image-text coherence but remain limited in capturing domain-specific or task-specific effectiveness, particularly for marketplace applications \citep{radford2021clip,hessel2021clipscore}. Our framework also motivates a new utility-aware evaluation method that balance image-text coherence and consumer preference.

\subsection{Regularized Learning}
Our proposed utility-aware model integrates a utility term into the contrastive learning objective. In this way, the augmented utility term can be viewed as a regularization term on the CLIP score. Regularization is widely used in statistical machine learning and data analytics, where different regularization terms serve different purposes, such as promoting sparsity \citep{tibshirani1996regression} and improving adversarial robustness \citep{xie2025asymptotic}, and incorporating geometric structure \citep{belkin2006manifold}. 
In our setting, the utility term in the proposed framework can be interpreted as a utility-based regularizer that steers the learned representation of the contrastive learning toward demand-enhancing solutions.

Prior work has also examined how business goals can be incorporated into objective functions for marketing optimization \citep{Lu2025Targeted}. However, to our knowledge, this principle has not been explored in the domain of text-to-image generation.

\section{Methodology: Utility-Aware Multimodal Contrastive Learning}\label{sec:method}

In this section, we introduce our Utility-Aware multimodal contrastive learning framework.
First, we review the contrastive loss in the Contrastive Language-Image Pretraining (CLIP) \citep{radford2021clip} and show how it can be interpreted through a discrete-choice point of view. 
Second, we introduce a Utility-Aware contrastive loss where the demand-driven visual utility component is added as a regularization in the similarity score. Third, we describe how our proposed Utility-Aware contrastive loss is used to update the pretrained CLIP model, yielding the Utility-Aware CLIP.
\subsection{Review of Contrastive Loss}\label{review}

Suppose we have matched image--text pairs
$\{(v_i,t_i)\}_{i=1}^N.$
The CLIP model learns a shared image--text representation space by optimizing a contrastive objective over matched image--text pairs. This objective is bidirectional: it includes a text-to-image InfoNCE loss and an image-to-text InfoNCE loss.
For example, the text-to-image InfoNCE loss is defined as follows
\begin{align}
\mathcal{L}^{t\to v}(\theta)
&=
-\frac{1}{N}\sum_{i=1}^N
\log
\frac{
e^{\frac{s_\theta(v_i,t_i)}{\tau}}
}{
\sum_{k=1}^N e^{\frac{s_\theta(v_k,t_i)}{\tau}}
},
\label{eq:infonce_t2v}
\end{align}
where $\tau > 0$ is a temperature hyperparameter and \begin{equation}\label{similarity}s_{\theta}(v, t) = u_v(v)^\top u_t(t)\end{equation} represents the cosine similarity between the visual embedding $u_v(v)$ and text embedding $u_t(t)$. $\theta$ denotes the set of learnable parameters, including the weights of the image and text encoders \footnote{An encoder is a model component that turns raw input, such as text or an image, into a numerical representation that AI can understand. In simple terms, it compresses the meaning of the input into a “digital summary” that the model can compare, search, or use for prediction.} along with their respective projection layers. The image-to text InfoNCE loss is defined analogously.

In the training process of CLIP model, the objective function is the average of the text-to-image InfoNCE loss and the image-to-text InfoNCE loss.

\subsection{Connection with Discrete Choice}\label{connection} 

The InfoNCE loss introduced in Section \ref{review} can be formalized through a discrete-choice perspective.

Consider first the text-to-image direction. Fix a text anchor \(t_i\) and a choice set of \(N\) sampled image alternatives \(\{v_1,\dots,v_N\}\). For each alternative \(v_k\), define
\[
U_{ik}^v = V_{ik}^v + \varepsilon_{ik}^v, \qquad k=1,\dots,N,
\]
where \(V_{ik}^v\) is the systematic utility of matching \(v_k\) with \(t_i\), and \(\varepsilon_{ik}^v\) is an i.i.d.\ Type-I extreme value error term. The image with the highest utility is selected with the conditional logit probability
\[
\Pr \bigl(v_i \mid t_i,\{v_1,\dots,v_N\}\bigr)
=
\frac{\exp(V_{ii}^v)}
{\sum_{k=1}^N \exp(V_{ik}^v)}.
\]
Setting
$V_{ik}^v = s_\theta(v_k,t_i)/\tau$,
and taking the negative average log-likelihood over all text anchors recovers the text-to-image InfoNCE loss in \eqref{eq:infonce_t2v}. Thus, the text-to-image InfoNCE loss is exactly the negative log-likelihood of a conditional logit model defined on the sampled image choice set.

The same reasoning applies to the image-to-text direction. Notably, minimizing the bidirectional InfoNCE contrastive loss is equivalent to maximum likelihood estimation of conditional logit models over finite sampled choice sets. This connection motivates a utility-aware extension of InfoNCE. Since the index inside the logit exponential represents systematic utility, the standard CLIP InfoNCE loss in \eqref{eq:infonce_t2v} corresponds to a similarity-only utility specification. This may be restrictive when observable attributes also affect matching utility. We therefore introduce a utility-aware similarity score and the corresponding Utility-Aware InfoNCE loss in the next subsection.

\subsection{Utility-Aware InfoNCE Loss}\label{utilityawareloss}

We introduce the Utility-Aware contrastive InfoNCE objective in this subsection. Let \(v\) denote an image and \(t\) denote a text description. For an image--text pair \((v,t)\), let \(s_\theta(v,t)\) be the standard CLIP similarity score. 
To incorporate demand-driven visual information, let \(h_v(v)\) denote a scalar visual utility score of image \(v\). This score may summarize demand-driven visual attributes, such as aesthetics, uniqueness, brightness, colorfulness, sustainability cues, or other image features. In empirical applications, \(h_v(v)\) can be constructed from an estimated demand model, for example as a weighted index of visual features. We define the utility-aware similarity score as
\begin{equation}
\label{eq:US_def}
US_{\theta,w}(v,t)
=
\alpha_v h_v(v)+\beta_s s_{\theta}(v,t),
\end{equation}
where \(\alpha_v,\beta_s\) are preference weights on the visual utility and semantic similarity components. 
We also refer to this utility-aware similarity measure as the Utility-Aware CLIP score, which can be used to evaluate both the semantic alignment and demand-related utility of a generated image.

Given \(N\) matched image--text pairs \(\{(v_i,t_i)\}_{i=1}^N\), the utility-aware text-to-image InfoNCE loss is
\begin{equation}
\label{eq:Lua_t2i_exp}
\mathcal{L}_{UA}^{t\to v}(\theta,w)
=
-\frac{1}{N}\sum_{i=1}^N
\log
\frac{\exp\left(\frac{US_{\theta,w}(v_i,t_i)}{\tau}\right)}
{\sum_{k=1}^N \exp\left(\frac{US_{\theta,w}(v_k,t_i)}{\tau}\right)}.
\end{equation}

The Utility-Aware image-to-text InfoNCE loss is defined analogously. The final objective is the bidirectional Utility-Aware contrastive loss, given by the average of the text-to-image and image-to-text components.

The proposed Utility-Aware loss nests the standard InfoNCE as a special case. When \(\alpha_v=0\) and \(\beta_s=1\), and the losses reduce to the standard CLIP InfoNCE losses discussed in Section~\ref{review}. 
\subsection{Utility-Aware Contrastive Language-Image Pretraining}\label{uclipsub}

Building on the Utility-Aware InfoNCE loss defined in subsection \ref{utilityawareloss}, we could obtain the  {Utility-Aware CLIP}. The goal is to adapt CLIP's image-text representation space so that it preserves not only semantic alignment between images and text, but also visual variation that is essential to marketplace performance. More specifically, the Utility-Aware CLIP is obtained by starting from pretrained CLIP weights and updating the model by optimizing the Utility-Aware InfoNCE loss.

\begin{figure}[htbp]
  \centering
  \includegraphics[width=0.9\linewidth]{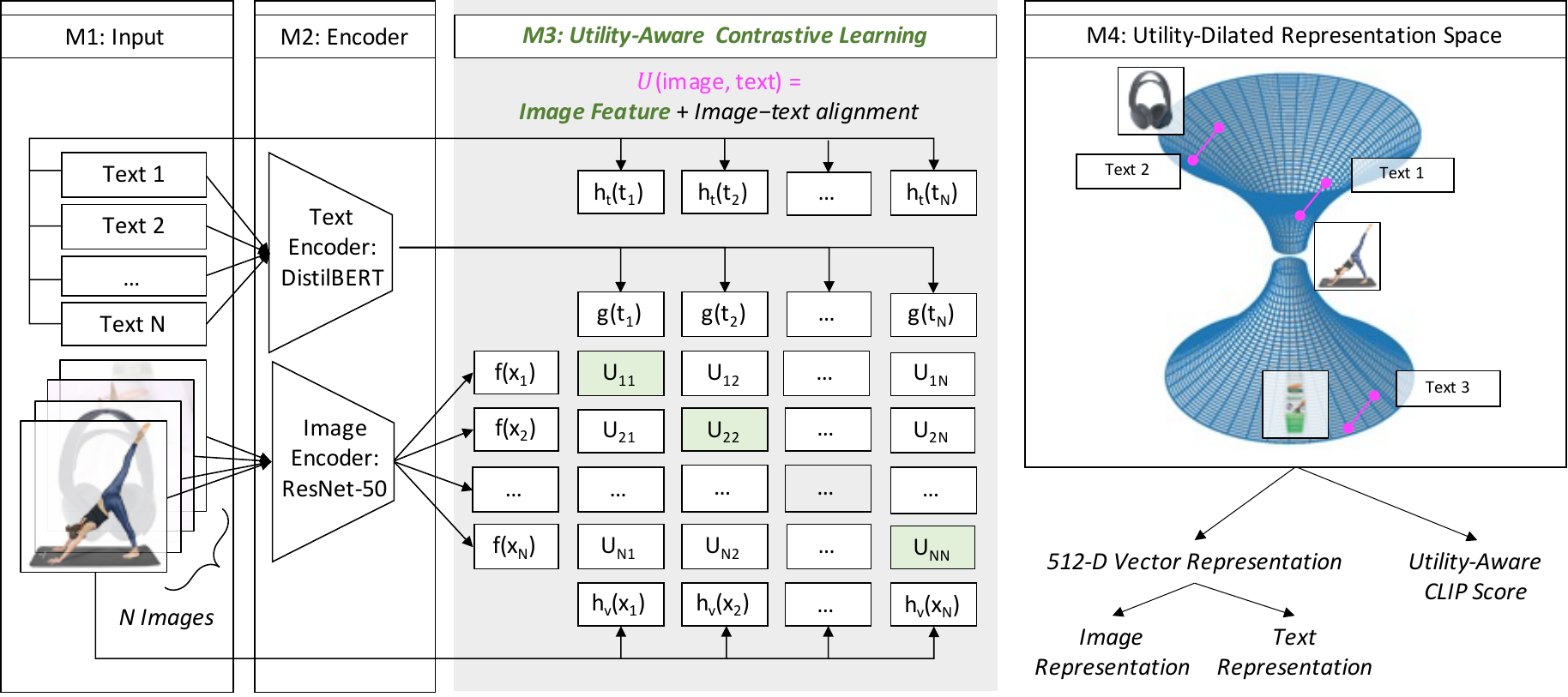}
  \caption{Utility-Aware Contrastive Image-Text Pretraining}
  \label{uclip_model_nsfc}
\end{figure}

The training process consists of four steps, as illustrated in Figure~\ref{uclip_model_nsfc}.

First, we construct positive and negative image--text pairs. For each product, the observed image \(v_i\) and its corresponding text \(t_i\) form a positive pair, while mismatched image--text combinations serve as negative pairs. 

Second, we encode images and text using the pretrained CLIP. The resulting embeddings are projected into a shared space, where image--text similarity is measured by cosine similarity. This follows the standard contrastive learning setup.

Third, we compute the Utility-Aware InfoNCE loss, which regularizes standard image--text similarity with a visual utility component based on demand-driven image features. This objective pulls semantically aligned and utility-driven matched pairs closer while separating mismatched or lower-utility alternatives.

Finally, we update the CLIP parameters based on the Utility-Aware InfoNCE loss. Starting from pretrained CLIP weights, this updates the image--text representation without training from scratch, yielding embeddings that preserve semantic alignment while emphasizing demand-driven visual attributes.

In our subsequent empirical analysis, we will apply the Utility-Aware CLIP to Amazon and Airbnb. The demand regularizer varies across different platforms: in the Amazon context, it is derived from visual attributes associated with product demand, while in the Airbnb context, it is derived from listing-level attributes associated with booking outcomes.We will discuss more details in Section \ref{sec:empirical}.

\section{Interpretation: Theoretical Properties and Interpretability}\label{interpretation}

The previous section defines the Utility-Aware InfoNCE loss and the Utility-Aware CLIP. We now explain why our utility-aware framework works. The Utility-Aware InfoNCE loss preserves CLIP's contrastive structure but compares candidate images using both semantic similarity and demand-driven visual utility. We study this interpretation through two components: theoretical properties of the loss and occlusion-based interpretation of the learned representation.

\subsection{Property of Utility-Aware InfoNCE Loss}\label{sec:mathprops}
In this subsection, we will provide a bound for the Utility-Aware  InfoNCE in terms of the mutual information.  Equipped with this bound, we could understand the theoretical interpretation and associated economic intuition of the regularization introduced in our utility-aware framework.

We  provide the bound of the proposed Utility-Aware InfoNCE in terms of the mutual information in the following theorem.

\begin{theorem}[Bound for Utility-Aware InfoNCE]\label{boundtheorem}
For the Utility-Aware InfoNCE defined in \eqref{eq:Lua_t2i_exp}, suppose $\tau=\beta_s=1$, then we have the following bound:
\begin{equation}\begin{aligned}\label{bound}&\log N-\mathcal{L}_{UA}^{t\to v}(\theta,w)\\
    \leq&   
   I(t,v) +\alpha_v \left(\frac{1}{N}\sum_{i=1}^N
 h_v(v_i)-\min_{1\leq i\leq N} h_v(v_i)\right),\end{aligned}\end{equation}
where $I(t,v)$ is the mutual information.\end{theorem}

The bound shown in Theorem~\ref{boundtheorem} is explicitly utility-aware. 
The contrastive bound for original InfoNCE in CLIP is only governed by the mutual information between image and text \citep{oord2018representation}. However, the bound for proposed Utility-Aware InfoNCE \eqref{bound} includes an additional term related to visual attribute \(h_v(v)\). Thus, utility-aware objective is able to capture both image--text alignment and the demand-driven image features, such as clarity, attractiveness, trustworthiness, informativeness, and compliance with platform standards. The coefficient \(\alpha_v\) controls the strength of additional utility component relative to semantic image--text matching.

Interestingly, our theory suggests that utility-aware training benefits from training data with heterogeneous visual attributes. This is consistent with the uniformity perspective of \cite{wang2020understanding}, which emphasizes that contrastive learning requires alternatives to be sufficiently spread out.
Our utility-aware framework indicates that, for commercial applications, achieving this spread requires a training dataset with not only semantic diversity across image-text pairs, but also variation in their demand-driven visual quality.

\subsection{Interpretability of the Representation Learned by Utility-Aware CLIP}

This subsection examines how the representation space learned by Utility-Aware CLIP differs from the representation space learned by standard CLIP.

We conduct an occlusion-based sensitivity analysis. 
For each image $v$ and prompt $t$, we divide the image into local patches. 
For each patch $r$, we construct a masked image $v^{(-r)}$ by occluding $r$ and define its importance as
\[
\Delta_r = \text{Score}(v,t)-\text{Score}(v^{(-r)},t).
\]
For standard CLIP, we set
$\text{Score}_{\text{CLIP}}(v,t)=s_{\theta}(v,t),$
whereas for Utility-Aware CLIP, we have
$
\text{Score}_{\text{UA}}(v,t)=US_{\theta,w}(v,t),$
where  $s_{\theta}(v,t)$ is the image-text similarity \eqref{similarity}, and $US_{\theta,w}(v,t)$ is the utility-regularized image-text similarity \eqref{eq:US_def}.  A larger $\Delta_r$ indicates that patch $r$ contributes more to the model score.

\begin{figure}[htbp]
\centering
\includegraphics[width=0.95\linewidth]{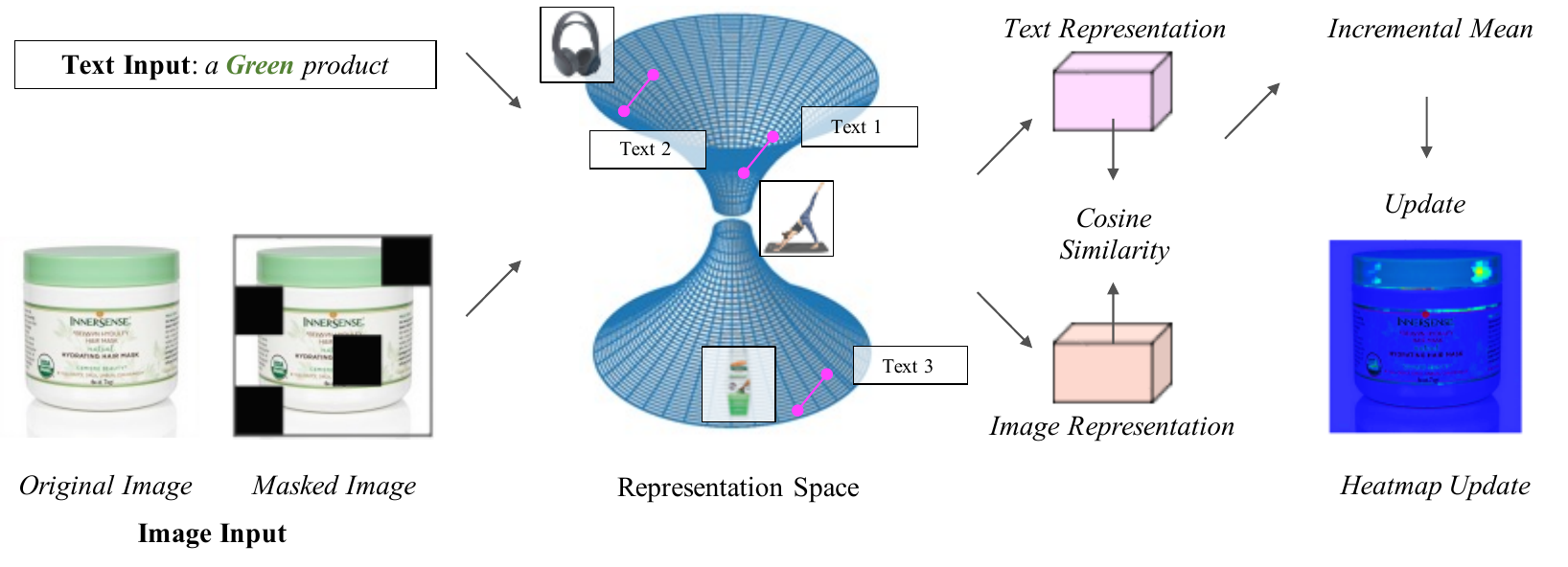}
\caption{Occlusion-Based Interpretation Framework}
\vspace{0.2cm}
\begin{minipage}{0.95\linewidth}
\small  {Note}: The image is masked patch by patch. For each masked image, we recompute the model score and measure the score drop relative to the original image. For the  CLIP model, the score is image-text similarity. For Utility-Aware CLIP, the score combines image-text similarity with demand-driven visual utility. Warmer regions indicate image patches that contribute more strongly to the model score.
\end{minipage}
\label{fig:fig14}
\end{figure}

\begin{figure}[htbp]
\centering
\subfloat[Table example: U-CLIP shows stronger and more spatially varied sensitivity. \label{fig:fig15a}]{
    \includegraphics[width=0.8\linewidth]{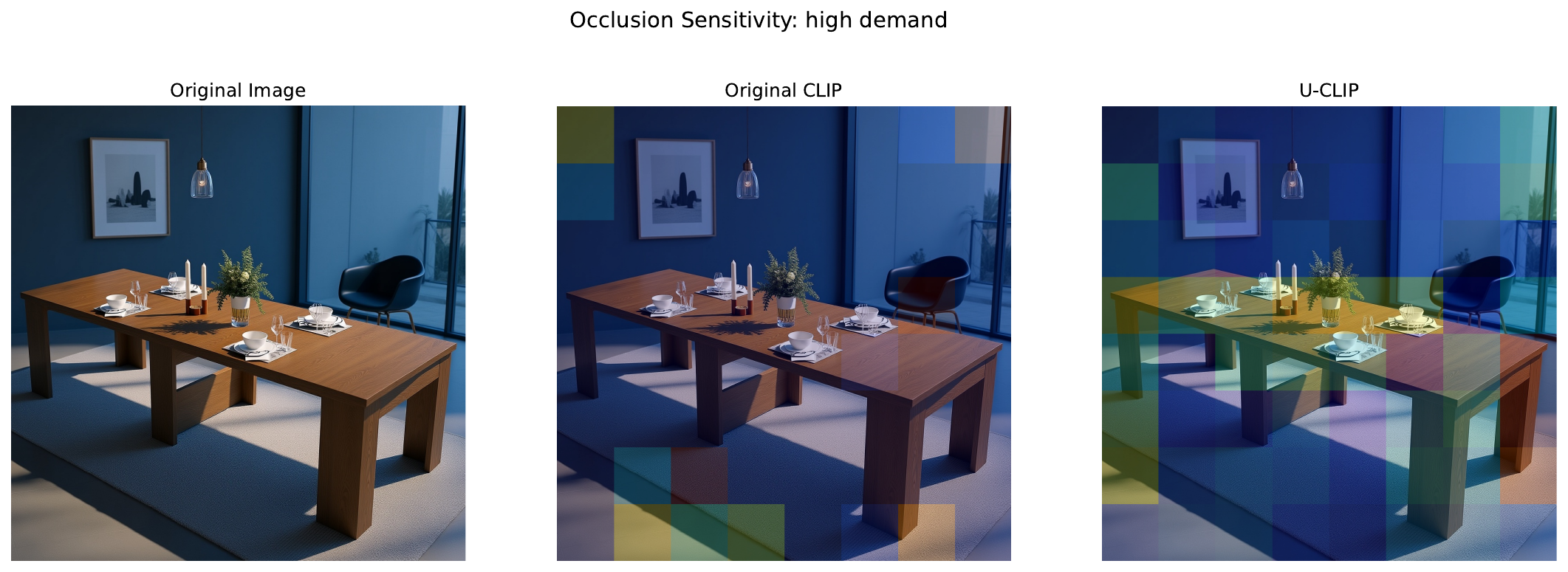}
} \\
\subfloat[Lamp example: U-CLIP captures more fine-grained demand-driven cues.\label{fig:fig15c}]{
    \includegraphics[width=0.8\linewidth]{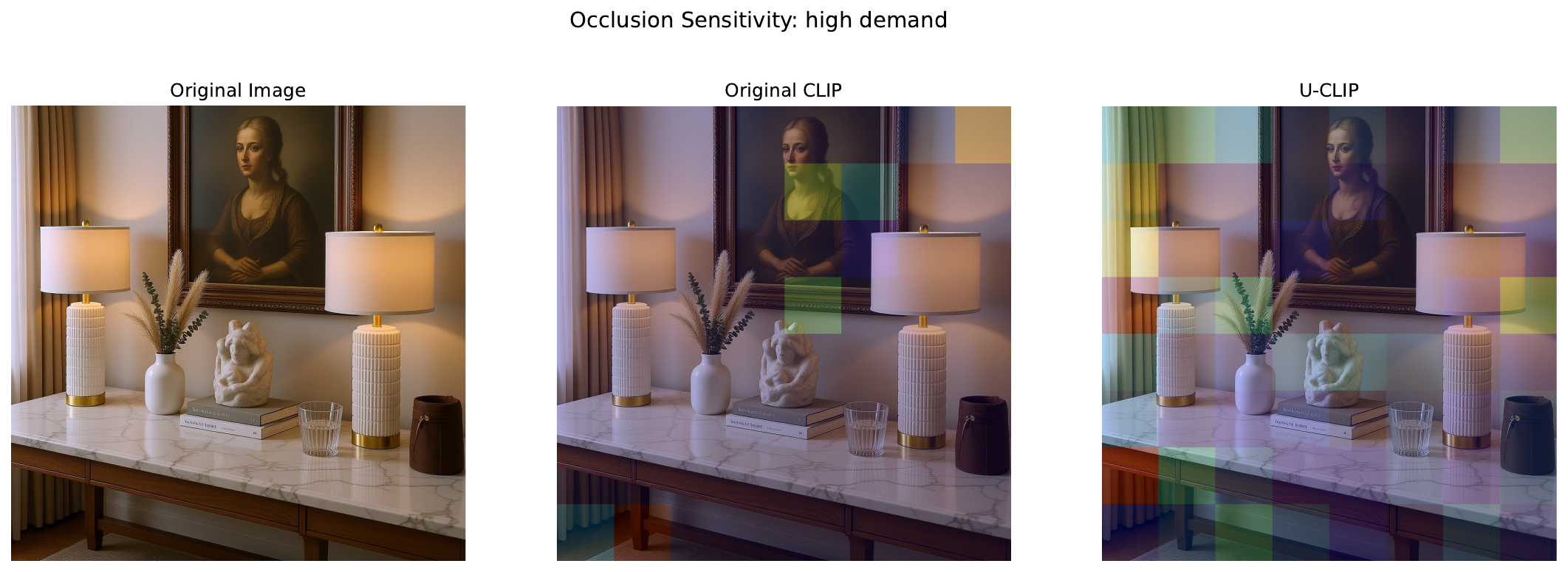}
}
\caption{Occlusion Interpretation of ``High Demand'' for Generated Amazon Product Images}
\vspace{0.5em}
\begin{minipage}{0.8\linewidth}
\small  {Note}: Each example shows the CLIP heatmap, and Utility-Aware CLIP heatmap for the prompt  {``high demand.''} Warmer colors indicate greater importance. Compared with CLIP, Utility-Aware CLIP produces more spatially differentiated heatmaps and is sensitive to broader details.
\end{minipage}
\label{fig:fig15}
\end{figure}

Figure~\ref{fig:fig14} illustrates the occlusion procedure and the construction of the resulting sensitivity heatmap.
Figure~\ref{fig:fig15} reports examples where we use the ``high demand'' as the prompt. The images are generated Amazon product images for tables and lamps. Each row compares the original image, the  CLIP occlusion heatmap, and the Utility-Aware CLIP occlusion heatmap. 
 
The comparison shows that Utility-Aware CLIP produces more spatially differentiated and interpretable sensitivity patterns than CLIP. While the CLIP responds mainly to regions supporting general image-text consistency, such as the main object or other salient areas, the Utility-Aware CLIP is sensitive to a broader set of demand-driven scene details, including product placement, lighting, surrounding furniture, tabletop arrangement, background composition, and lifestyle context.

This pattern is consistent with what makes product images effective in marketplace settings. High-demand images depend not only on the product itself, but also on staging, realism, lighting, and a credible usage context.
The greater spatial variation in the Utility-Aware CLIP heatmaps suggests that the model has learned a more discriminating visual representation: some regions receive high sensitivity, while others receive low sensitivity. Rather than treating the image only as a generic match to the prompt, Utility-Aware CLIP identifies which components of the product scene contribute more strongly to the high-demand signal.

Overall, the occlusion analysis suggests that the utility-aware objective changes how the model interprets image-text alignment. While CLIP primarily captures semantic consistency with the prompt, Utility-Aware CLIP preserves this connection while shifting sensitivity toward regions that are more relevant for marketplace evaluation. This provides an interpretable link between the demand-driven training objective, the learned multimodal representation, and the generation of product images visually aligned with commercial performance.

\section{Application: Utility-Aware Product Image Generation}\label{sec:empirical}

In this section, we illustrate the application of the proposed Utility-Aware multimodal contrastive learning introduced in Section
\ref{sec:method}. 
More specifically, we discuss the resulting Utility-Aware Generator and its applications on Amazon and Airbnb, using context-specific demand models to promote demand-related visual attributes without producing overly stylized or unrealistic images.

\subsection{Utility-Aware Generator}

We embed the Utility-Aware CLIP from Section \ref{sec:method} into the state-of-the-art open-source image generation model, Flux \citep{flux-2-2025}, as the Utility-Aware Generator. The key innovation is replacing Flux's standard CLIP encoder with our Utility-Aware CLIP encoder.

The Utility-Aware Generator accepts two inputs: a text prompt and an optional reference image. The baseline Flux model generates candidate images conditioned on these inputs. We replace the original CLIP encoder in the Flux with our Utility-Aware CLIP encoder, which evaluates generated images on both semantic consistency with the prompt and alignment with business-relevant visual attributes, e.g., sustainability cues, product fidelity, aesthetics, realism. This substitution adds a demand-guided layer to the Flux generation process without altering the core generative architecture.

After generation, we perform post-hoc selection of candidate images based on Utility-Aware CLIP scores as defined in \eqref{eq:US_def}. The model generates multiple image samples, and we rank them using the Utility-Aware CLIP score, which is a weighted combination of semantic similarity and estimated demand utility. This selection step ensures that the final output not only preserves fidelity to the input prompt but also reflects visual attributes associated with higher demand in the target context.

We instantiate the Utility-Aware Generator in Amazon and Airbnb settings by adjusting the demand regularizer to match each context. We consider two tasks: image generation (producing new images from text prompts) and image editing (improving visual quality or relevance while preserving original design). Because editing  better reflects real-world business use cases, it serves as our main focus, while pure text-to-image generation results are reported in the Appendix. Our model generation output is proved to be robust across different prompt variations.

\begin{figure}[htbp]
  \centering
  \includegraphics[width=1.0\linewidth]{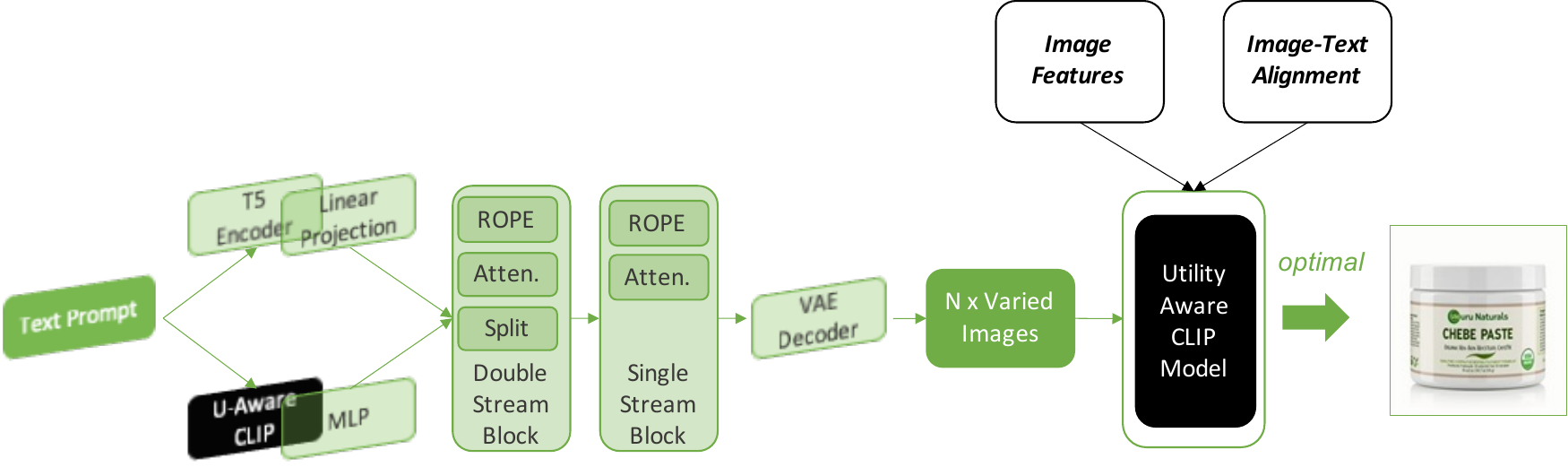}
  \caption{Utility-Aware Generation Framework: Flux baseline with Utility-Aware CLIP encoder substitution and post-hoc image selection}
  \label{uclip_framework}
\end{figure}

\subsection{Application I: Amazon}

We try to edit existing product images to create additional images showing possible usage scenarios. The goal is to create realistic in-context images that could be added to Amazon listings, without changing the focal product or incurring additional photography and design costs. To guide this editing task, we calibrate the utility-aware objective using observed consumer behavior on Amazon.
\subsubsection{Demand Model Estimation}
To quantify how visual attributes affect product performance, we estimate a reduced-form regression where the dependent variable is log sales rank, which serves as an inverse demand proxy. The key explanatory variables are visual attributes, including colorfulness, brightness, symmetry, and aesthetic quality. Before estimation, we rescale all visual attributes to lie in the $[0,1]$ interval. This normalization ensures comparability across features and preserves the curvature implied by the quadratic terms. We also include sub-category fixed effects, firm size controls, and month fixed effects:

\begin{equation*}
\begin{aligned}
\text{Demand}_{i} =
&\beta_{\text{c1}} C(v_i) + \beta_{\text{c2}} C(v_i)^2 \\
+& \beta_{\text{b1}} B(v_i) + \beta_{\text{b2}} B(v_i)^2 \\
+& \beta_{\text{s1}} S(v_i) + \beta_{\text{s2}} S(v_i)^2 \\
+& \beta_{\text{a1}} A(v_i) + \beta_{\text{a2}} A(v_i)^2 \\
+& \gamma \mathbf{X}_i + \varepsilon_i,
\end{aligned}
\end{equation*}
where $C(v)$, $B(v)$, $S(v)$, and $A(v)$ denote colorfulness, brightness, symmetry, and aesthetic quality, respectively, and $\mathbf{X}_i$ includes sub-category, firm size, and time fixed effects.

Table~\ref{tab:visual_demand} reports the estimation results. We find strong and consistent evidence of non-linear effects across major visual dimensions. In particular, colorfulness has a negative linear coefficient ($\beta_{\text{c1}}=-2.249$) and a positive quadratic coefficient ($\beta_{\text{c2}}=2.118$). Because log sales rank is an inverse demand proxy, this pattern implies that demand is highest at intermediate levels of colorfulness and lower at extreme levels. A similar pattern emerges for brightness, symmetry, and aesthetic quality. For example, aesthetic quality shows one of the strongest non-linear effects ($\beta_{\text{a1}}=-2.885$, $\beta_{\text{a2}}=2.951$), suggesting that excessive stylization may reduce perceived authenticity or relevance.

\begin{table}[htbp]
\centering
\caption{Visual Attributes and Demand on Amazon}
\label{tab:visual_demand}
\begin{tabular}{lcc}
\toprule
D.V.: Log(Sales Rank) & Coefficient & Std. Error \\
\midrule
Colorfulness & -2.249*** & (0.075) \\
Colorfulness$^2$ & 2.118*** & (0.073) \\

Brightness & -0.281** & (0.087) \\
Brightness$^2$ & 0.588*** & (0.091) \\

Symmetry & -1.074*** & (0.104) \\
Symmetry$^2$ & 1.664*** & (0.162) \\

Aesthetic Quality & -2.885*** & (0.087) \\
Aesthetic Quality$^2$ & 2.951*** & (0.077) \\

\midrule
Sub-category FE & Yes &  \\
Firm Size Controls & Yes &  \\
Month FE & Yes &  \\

Observations & 341,779 &  \\
$R^2$ & 0.440 &  \\
\bottomrule
\end{tabular}

\vspace{2mm}
\begin{flushleft}
\footnotesize
 {Notes:} The dependent variable is a proxy for demand, log(sales rank), which is the inverse of sales volume. All visual attributes are standardized. Standard errors are reported in parentheses. *** $p<0.01$, ** $p<0.05$, * $p<0.1$.
\end{flushleft}
\end{table}
\subsubsection{Utility-Aware Generator Training}
We incorporate the estimated demand function into the utility-regularized similarity function as:

\begin{equation*}
\begin{aligned}\label{utilityregamazon}
US(v,t) &= \eta \Big(
\beta_{\text{c1}} C(v) + \beta_{\text{c2}} C(v)^2 \\
&\quad + \beta_{\text{b1}} B(v) + \beta_{\text{b2}} B(v)^2 \\
&\quad + \beta_{\text{s1}} S(v) + \beta_{\text{s2}} S(v)^2 \\
&\quad + \beta_{\text{a1}} A(v) + \beta_{\text{a2}} A(v)^2
\Big) \\
&\quad + s_{\theta}(v,t),
\end{aligned}
\end{equation*}
where $s_{\theta}(v,t)$ denotes the cosine similarity between the image and text embeddings, and $\eta$ is a hyperparameter controlling the strength of the utility-aware regularization. By tuning $\eta$, practitioners can adjust the relative weight placed on semantic similarity and demand-related visual attributes.
This formulation  embeds the nonlinear preference structure implied by observed demand directly into the similarity metric. The quadratic terms allow the model to favor intermediate, demand-maximizing levels of visual attributes instead of pushing each attribute higher.

We train the Utility-Aware CLIP model on the 7,719 matched Amazon product images and descriptions following Section \ref{sec:method} based on the utility-regularized similarity function \eqref{utilityregamazon}.
 
\subsubsection{Performance Evaluation}
We then generate additional usage-scenario images for existing products, such as lamps, blankets, and tables, using Stable Diffusion, GPT-Image, Flux, and the Utility-Aware Generator. Generated images are evaluated using: (i) a demand score, (ii) similarity to the original product, i.e., fidelity, (iii) Utility-Aware CLIP  (U-CLIP) score, (iv) aesthetic quality, and (v) visual attributes such as brightness and colorfulness. 

Our evaluation framework isolates the  contribution of visual attributes to product demand by holding all non-visual product features constant. The \textit{demand score} is computed by applying the estimated demand function as shown in Table~\ref{tab:visual_demand} to each generated image, allowing only the visual features including colorfulness, brightness, symmetry, and aesthetic quality, to vary while fixing product characteristics such as category, firm size, and time period at their baseline levels. This approach directly measures whether our Utility-Aware Generator produces images that align with the revealed preference structure learned from real sales data, without confounding effects from product attributes outside the scope of image editing. The \textit{fidelity} score measures visual and semantic consistency with the original product image, ensuring that demand improvements do not come at the cost of misrepresenting the actual product. The \textit{Utility-Aware CLIP} score captures how well the learned representation space shaped by demand data aligns with the generated image and text. The individual visual attributes such as brightness, colorfulness, and aesthetic quality are reported to verify that our generator balances these attributes at their demand-enhancing  levels rather than pushing any attribute to extremes.
This is consistent with the regression results, which show nonlinear preferences with an inverted U shape.
These metrics can help validate that the Utility-Aware Generator improves sales potential through economically meaningful visual refinements while preserving product authenticity and operational feasibility for e-commerce platforms.

\begin{table}[htbp]
\centering
\caption{Comparison among Generative Models in Image Editing Task (Amazon Context)}
\label{tab:comparison-models}
\renewcommand{\arraystretch}{1.2}
\setlength{\tabcolsep}{5pt}
\begin{tabular}{lcccccc}
\hline
\textbf{Model} & \textbf{Demand} & \textbf{Fidelity} & \textbf{U-CLIP} & \textbf{Aesthetic} & \textbf{Brightness} & \textbf{Colorfulness} \\
\hline
Stable Diffusion & 1.149 & 0.224 & 0.270 & 0.706 & 0.270 & 0.306 \\
GPT-Image  & 1.007 & 0.187 & 0.322 & 0.686 & 0.402 & 0.426 \\
Flux & 1.159 & 0.228 & 0.226 & 0.772 & 0.144 & 0.440 \\
\textit{\shortstack[l]{Utility-Aware\\Generator}}&  \textit{1.278} &  \textit{0.231} &  \textit{0.443} &  \textit{0.713} &  \textit{0.177} &  \textit{0.414} \\
\hline
\end{tabular}

\vspace{2mm}
\begin{flushleft}
\footnotesize
 {Notes:} Demand is computed using the estimated demand function combining similarity and image features with non-linear transformations. Fidelity corresponds to similarity to the original listing image. Brightness, colorfulness, and aesthetic scores are rescaled components entering the demand function.
\end{flushleft}
\end{table}

Table~\ref{tab:comparison-models} reports average performance across 100 products. Consistent with the non-linear demand estimates, baseline generative models tend to produce more extreme visual characteristics. For example, GPT-Image generates relatively high brightness and colorfulness, while Flux produces highly stylized scenes with elevated aesthetic scores. These shifts do not consistently translate into higher demand, reflecting the non-monotonic preference structure identified in the regression.

In contrast, the Utility-Aware Generator achieves the highest demand score, outperforming all baselines. This improvement arises because the model internalizes the estimated curvature of the demand function and generates usage scenarios that balance visual attributes rather than pushing them to extremes. The resulting images exhibit moderate brightness and colorfulness, while also maintaining the highest fidelity to the original product. Overall, these results suggest that incorporating empirically estimated, non-linear visual preferences is valuable for product image redesign on Amazon. By avoiding over-optimization of individual attributes and targeting balanced, realistic contexts, the Utility-Aware Generator produces images that are both economically effective and operationally scalable for e-commerce platforms.

\begin{figure}[htbp]
\centering
    \includegraphics[width=0.8\linewidth]{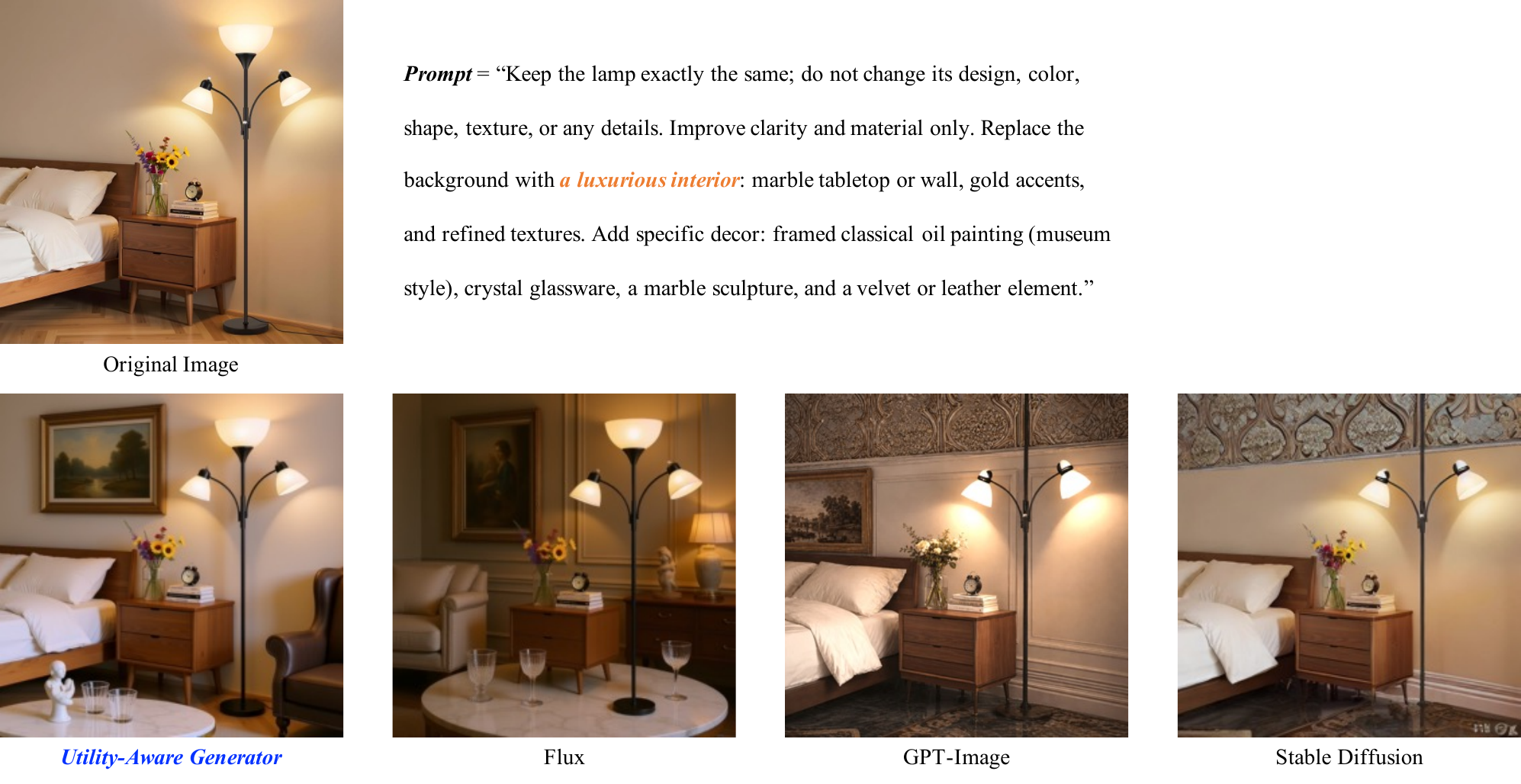}
    \caption{Amazon Product Image Editing (Example 1)}
    \label{fig:amazon_edit_exp1}
\end{figure}

\begin{figure}[h]
    \centering
    \includegraphics[width=0.8\linewidth]{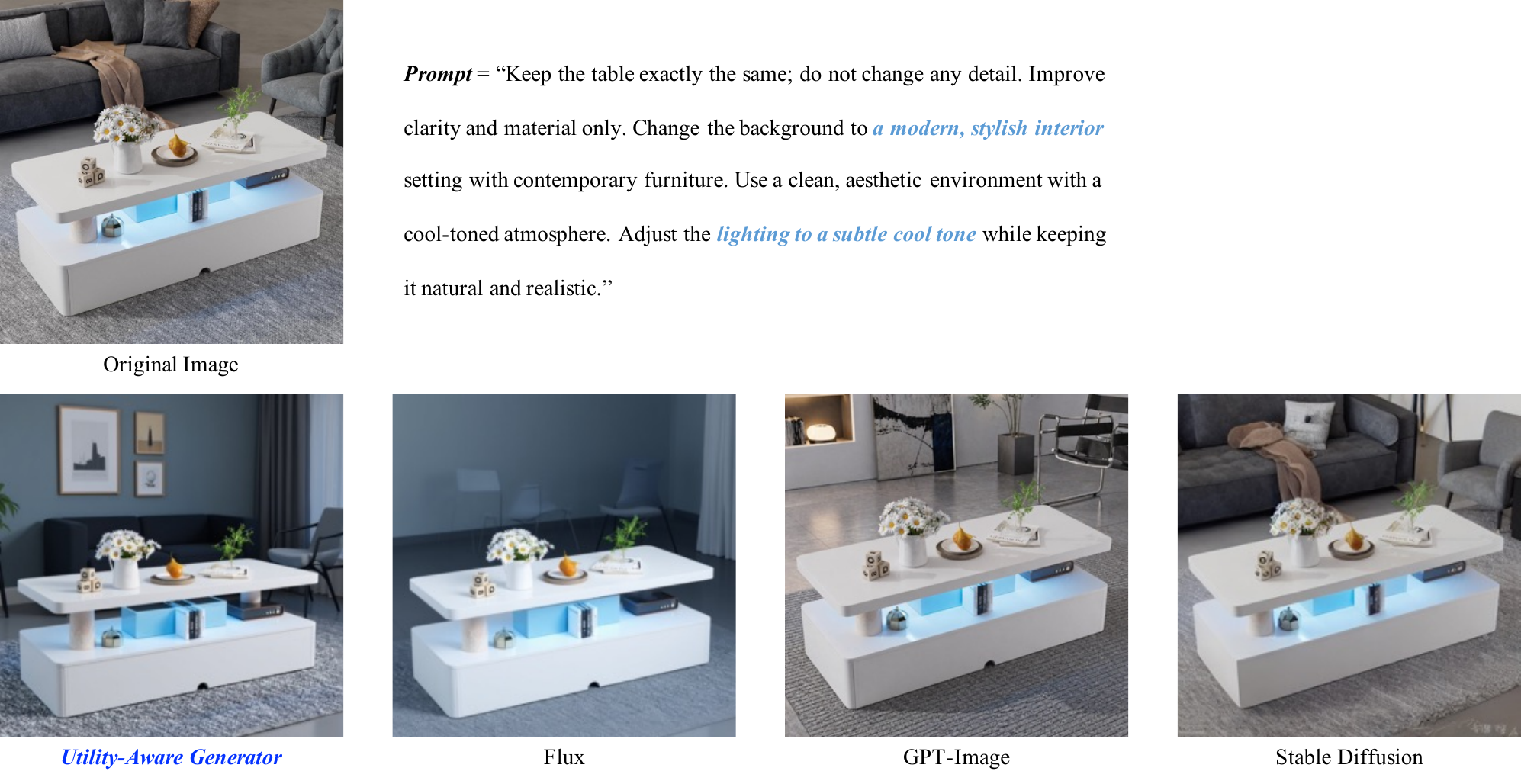}
    \caption{Amazon Product Image Editing (Example 2)}
    \label{fig:amazon_edit_exp2}
\end{figure}

As shown in Example~1 in Figure~\ref{fig:amazon_edit_exp1}, the editing process improves clarity and perceived material quality while maintaining fidelity to the original product. The generated scene introduces subtle decorative elements and adjusts brightness, aesthetics, and colorfulness to moderate levels, avoiding over enhancement while keeping the product as the focal point. Example~2 in Figure~\ref{fig:amazon_edit_exp2} enhances the product image with a modern interior setting while preserving the original product design. The background create a more refined usage scenario, but the edits remain moderate and realistic.

\subsection{Application II: Airbnb}

In the Airbnb application, we study whether the proposed utility-aware framework can edit property images to increase demand. This setting differs from Amazon because the focal objective is not product fidelity, but listing realism and booking relevance. We therefore construct the utility-aware objective using Airbnb-specific visual attributes, especially visual uniqueness and aesthetic quality, while preserving semantic consistency with the listing description.

\subsubsection{Demand Model Estimation}
To capture the factors that drive engagement in this setting, 
we estimate a demand model for booking outcomes. The dependent variable is log occupancy rate. The key explanatory variables are visual uniqueness and aesthetic quality, each entered with both linear and quadratic terms, alongside standard controls:

\begin{equation*}
\begin{aligned}
\text{Demand}_{i} =
&\beta_{\text{u1}} U(v_i) + \beta_{\text{u2}} U(v_i)^2\\
+& \beta_{\text{a1}} A(v_i) + \beta_{\text{a2}} A(v_i)^2\\
+& \gamma \mathbf{X}_i + \varepsilon_i,
\end{aligned}
\end{equation*}
where uniqueness score $U(v)$ is computed using a self-supervised visual representation model, encouraging images that deviate from common patterns in the training distribution \citep{feng2025uniqueness}, the aesthetic score $A(v)$ is obtained using the Neural Image Assessment (NIMA) model \citep{talebi2018nima}, which evaluates perceptual quality and visual attractiveness, and $\mathbf{X}_i$ includes log nightly price, listing characteristics and location-time fixed effects.

Table~\ref{tab:visual_demand} reports the estimation results. We find strong evidence of an inverted U-shaped relationship between these visual attributes and demand: moderate levels of uniqueness and aesthetic quality maximize booking outcomes, while excessively low or high levels reduce demand.

\begin{table}[htbp]
\centering
\caption{Visual Attributes and Demand on Airbnb}
\label{tab:demand2}
\begin{tabular}{lcc}
\toprule
D.V.: Log(Occupancy Rate) & Coefficient & Std. Error \\
\midrule
Visual Uniqueness & 0.288*** & (0.0456) \\
Visual Uniqueness$^2$ & -0.241*** & (0.0364) \\
Visual Aesthetics & 0.978*** & (0.0656) \\
Visual Aesthetics$^2$ & -1.013*** & (0.0663) \\
Property Daily Rate & -0.158*** & (0.0114) \\
Response Rate & 0.0532*** & (0.0040) \\
Overall Rating & 0.324*** & (0.0161) \\
Visual Complexity & -0.237*** & (0.0325) \\
Number of Images & 0.0130*** & (0.00145) \\
Number of Reviews & 0.0219*** & (0.00103) \\
Airbnb Superhost & 0.00495*** & (0.00135) \\
Number of Max Guests & 0.101*** & (0.00953) \\
Instant Book Enabled & -0.0154*** & (0.00133) \\
Minimum Stay & -1.80e-05 & (4.08e-05) \\
Number of Room Amenities & 0.0323*** & (0.00224) \\
Bedrooms & 0.0101*** & (0.00128) \\
Bathrooms & 0.0189*** & (0.00144) \\
Median Age & 0.000468** & (0.000197) \\
Median Income & -1.21e-06*** & (3.90e-08) \\
Children & 0.133*** & (0.0345) \\
Married & -0.0335 & (0.0208) \\
Bachelors & 0.416*** & (0.0126) \\
Supply Count & 0.00401*** & (0.000819) \\
Constant & 0.192*** & (0.0266) \\
\midrule
County-Month FE & Yes &  \\
Observations & 95,673 &  \\
$R^2$ & 0.251 &  \\
\bottomrule
\end{tabular}
\vspace{2mm}
\begin{flushleft}
\footnotesize
{Notes:} The dependent variable is a proxy for demand, log(occupancy rate), which is the proportion of days that a property was booked in a month. All visual attributes are standardized. Standard errors are reported in parentheses. *** $p<0.01$, ** $p<0.05$, * $p<0.1$.
\end{flushleft}
\end{table}

\subsubsection{Utility-Aware Generator Training}
To make the demand estimation directly usable in our model, we rescale uniqueness and aesthetics to the $[0,1]$ interval. This normalization places both attributes on a common scale and allows the estimated coefficients to be incorporated into the learning objective without additional transformation:

\begin{equation*}
\begin{aligned}
US(v,t) &= \eta \Big(
\beta_{\text{u1}} U(v) + \beta_{\text{u2}} U(v)^2 \\
&\quad + \beta_{\text{a1}} A(v) + \beta_{\text{a2}} A(v)^2
\Big) \\
&\quad + s_{\theta}(v,t),
\end{aligned}
\end{equation*}
where $s_{\theta}(v,t)$ denotes the similarity, and $\eta$  controls the strength of the utility-aware regularization.
This formulation embeds the nonlinear preference structure from the demand data directly into the similarity metric. The quadratic terms allow the model to favor intermediate, demand-maximizing levels of uniqueness and aesthetics rather than mechanically pushing these attributes higher.

We train the Utility-Aware CLIP model on 10,000 Airbnb property listings, each with a cover image and a corresponding textual description. The cover image is the primary visual signal shown in search results and critically influences user attention and perceived property quality.
\subsubsection{Performance Evaluation}
To evaluate generative performance of the Utility-Aware Generator, we sample 500 bedroom descriptions from the dataset and generate images using four models: Stable Diffusion, DALL$\cdot$E, Flux, and our Utility-Aware Generator. Unlike baseline models that primarily optimize for text-image correspondence, our generator uses the learned utility-aware representation to guide generation toward images that are semantically aligned, visually engaging, and consistent with the estimated demand function. 

Our evaluation framework isolates the contribution of visual attributes to booking demand by holding all non-visual listing features constant. The \textit{demand score} is computed by applying the estimated demand function in Table~\ref{tab:demand2} to each generated image, allowing only the visual features including uniqueness and aesthetic quality to vary while fixing property characteristics including price, amenities, location, and host rating at their sample means. This approach directly measures whether our Utility-Aware Generator produces images that align with the revealed preference structure learned from real booking data, without confounding effects from listing attributes outside the scope of image generation. The \textit{fidelity} score measures semantic and visual consistency with the original property image, ensuring that demand improvements do not come at the cost of misleading guests about actual property appearance. The \textit{uniqueness} and \textit{aesthetic} scores capture the individual components of the demand function, allowing us to verify that our generator balances these attributes at their demand-maximizing levels rather than pushing either to extremes. Together, these metrics will help validate that the Utility-Aware Generator improves booking potential through economically meaningful visual adjustments while preserving realism and property authenticity (Table~\ref{tab:comparison-models-t2i}). Detailed analysis and example images are in the Appendix. 

\begin{table}[htbp]
\centering
\caption{Comparison among Generative Models in Image Generation Task (Airbnb Context)}
\label{tab:comparison-models-t2i}
\renewcommand{\arraystretch}{1.2}
\setlength{\tabcolsep}{6pt}
\begin{tabular}{lcccc}
\hline
\textbf{Model} & \textbf{Demand} & \textbf{Fidelity} & \textbf{Uniqueness} & \textbf{Aesthetic} \\
\hline
Stable Diffusion & 0.505 & 0.213 & 0.556 & 0.642 \\
DALL$\cdot$E     & 0.510 & 0.224 & 0.551 & 0.660 \\
Flux             & 0.544 & 0.251 & 0.549 & 0.641 \\
 \textit{Utility-Aware Generator} &   \textit{0.573} &   \textit{0.268} &   \textit{0.547} &   \textit{0.602} \\
\hline
\end{tabular}

\vspace{2mm}
\begin{flushleft}
\footnotesize
 {Notes:} Demand is computed using the estimated demand function combining similarity, uniqueness, and aesthetic quality with non-linear transformations. Fidelity corresponds to similarity to the original listing image. Uniqueness and aesthetic scores are rescaled components entering the demand function.
\end{flushleft}
\end{table}

We next consider a more practical task: improving existing property images through controlled editing. The instruction is fixed across images and intentionally simple. Its role is to specify a general direction for plausible visual refinement, such as adding decorative elements and soft furnishings. Specifically, the prompt asks the model to improve room aesthetics and uniqueness by adding two pictures, a carpet, pillows, and flowers, with the goal of increasing the occupancy rate. The desired style is clean, cozy, warm, and artistic. Importantly, the instruction does not determine the optimal degree of visual change. Rather, editing is carried out in a representation space shaped by demand-driven preferences learned from the data.

In image editing, visual changes are generated under our utility-aware framework, so edits follow learned preference structures without additional prompt engineering or test-time optimization. This typically yields moderate, coherent modifications, e.g., artwork, textiles, decorative accents, while preserving layout and realism.

Table~\ref{tab:comparison-models-airbnb} reports average performance across generated images.
\begin{table}[htbp]
\centering
\caption{Comparison among Generative Models in Image Editing Task (Airbnb Context)}
\label{tab:comparison-models-airbnb}
\renewcommand{\arraystretch}{1.2}
\setlength{\tabcolsep}{6pt}
\begin{tabular}{lcccc}
\hline
\textbf{Model} & \textbf{Demand} & \textbf{Fidelity} & \textbf{Uniqueness} & \textbf{Aesthetic} \\
\hline
Stable Diffusion & 0.505 & 0.206 & 0.566 & 0.609 \\
GPT-Image  & 0.508 & 0.214 & 0.554 & 0.637 \\
Flux & 0.507 & 0.209 & 0.565 & 0.618 \\
 \textit{Utility-Aware Generator} &  \textit{0.521} & \textit{0.225} &  \textit{0.527} &  \textit{0.623} \\
\hline
\end{tabular}

\vspace{2mm}
\begin{flushleft}
\footnotesize
 {Notes:} Demand is computed using the estimated demand function combining similarity, uniqueness, and aesthetic quality with non-linear transformations. Fidelity corresponds to similarity to the original listing image. Uniqueness and aesthetic scores are rescaled components entering the demand function.
\end{flushleft}
\end{table}
The Utility-Aware Generator achieves the highest demand score and the highest fidelity score. It also avoids pushing uniqueness to the highest level. This pattern is consistent with the inverted U-shaped demand relationship: the goal is not to maximize uniqueness, but to reach a balanced level that improves perceived appeal while preserving realism.

\begin{figure}[htbp]
\centering

    \centering
    \includegraphics[width=0.6\linewidth]{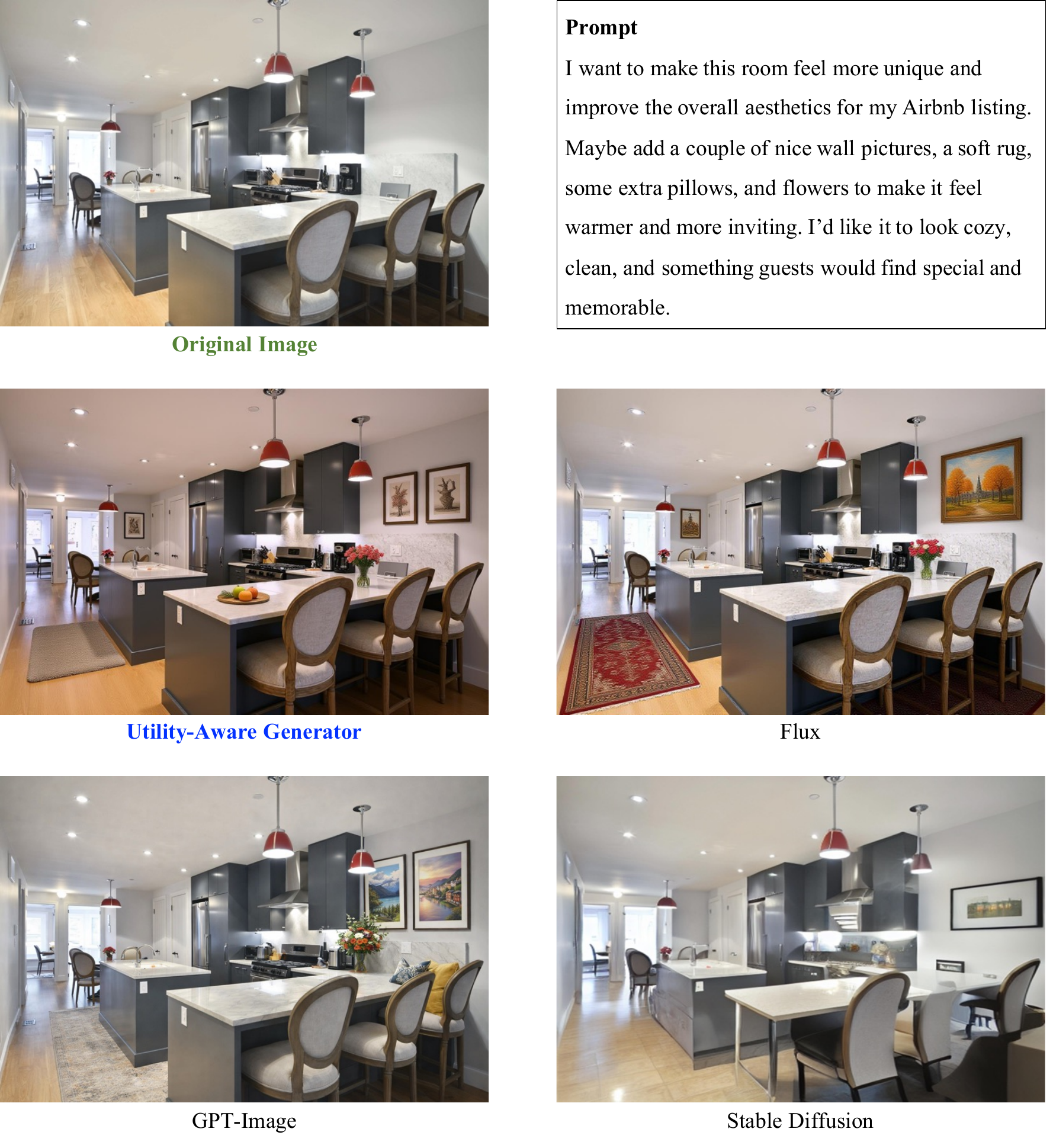}
    \caption{Kitchen example: utility-aware editing introduces moderate decorative elements, such as artwork, flowers, and textiles, that enhance warmth and distinctiveness while preserving realism.}
\end{figure}

\vspace{0.5cm}

\begin{figure}[h]
    \centering
    \includegraphics[width=0.6\linewidth]{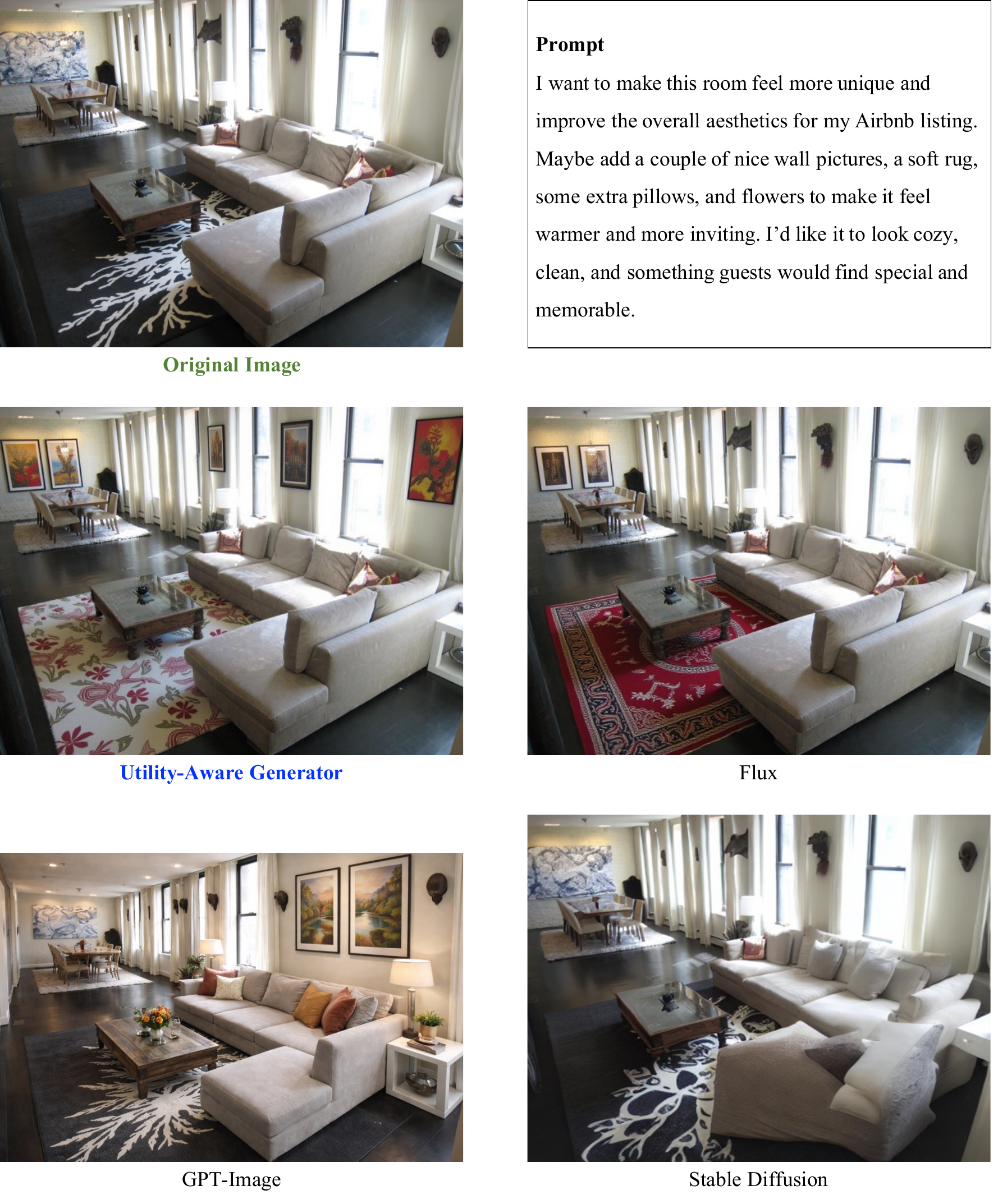}
    \caption{Living room example: utility-aware editing increases visual richness through color, soft furnishings, and wall art without generating excessive clutter or unrealistic changes.}
\end{figure}

\section{Validation: Human Experiment}
To evaluate the external validity of our approach, we conduct two randomized survey studies in distinct decision contexts: product evaluation and Airbnb booking. In both settings, participants are exposed to original images alongside AI-generated alternatives from multiple models with randomized positions, and are asked to evaluate or select among them. This design allows us to assess whether differences in generated visual quality translate into meaningful differences in human perception and choice. Across both studies, we combine model-free comparisons with regression analyses to isolate the impact of our model relative to competing approaches. Details of these surveys are in the Appendix. 

\subsection{Study I: Validating Utility-Aware Editing Product Images on Amazon}

We evaluate whether images generated by different models vary in perceived quality and consumer choice using a randomized choice-based survey. In each task, participants viewed an original product image alongside four AI-edited alternatives with randomized positions. Each participant was asked to complete three tasks: (i) select the image they would be most likely to purchase, (ii) rate each image on perceived realism and trustworthiness, and (iii) rate each image on professional quality for commercial use. We recruited 118 participants from Prolific.com. Two participants were excluded due to incomplete responses. Each participant completed six choice tasks, yielding 708 choice sets and 2,832 image-level observations. In each task, participants were asked to evaluate four AI-generated images produced by different models: Stable Diffusion, OpenAI, Flux, and the Utility-Aware Generator, with image-to-position assignment randomized.

We begin with model-free evidence. Table~\ref{tab:descriptive} reports raw selection counts and average ratings by model. The Utility-Aware Generator is selected substantially more often than the competing models and also receives the highest average ratings on realism and trustworthiness, and professional quality. These show that participants perceive images produced by the Utility-Aware Generator as more commercially appealing and more suitable for marketplace use.

\begin{table}[t]
\centering
\caption{Model-Free Differences Across Image Generation Models}
\label{tab:descriptive}
\begin{tabular}{lccc}
\toprule
Model & Selection Count & Realism \& Trustworthiness & Professional Quality \\
\midrule
Stable Diffusion & 154 & 4.54 & 4.55 \\
OpenAI & 143 & 4.64 & 4.74 \\
Flux & 59  & 4.03 & 4.15 \\
Utility-Aware Generator & 256 & 5.12 & 5.23 \\
\bottomrule
\end{tabular}\\
\vspace{0.5em}
\footnotesize{
Notes: Table reports raw (unadjusted) averages across all observations. 
Selection Count denotes the total number of times images from each model were chosen. 
Ratings are measured on 7-point Likert scales.
}
\end{table}

To formally estimate the effect of model type, we construct an image-level dataset and estimate the following models. We first examine how model type affects perceived image quality:

\begin{align*}
\text{Rate}_{ijkt} &= \beta_1 \text{UA}_k + \gamma P_k + \epsilon_{ijkt}
\end{align*}
where $\text{Rate}_{ijkt}$ denotes image $k$ rating by participant $i$ in choice set $j$. $\text{UA}_k=1$ indicates the image is generated by the Utility-Aware Generator and vice versa. $\beta_1$ captures the average difference in perceived quality between images generated by the Utility-Aware Generator and alternative models. $P_k$ denotes image position fixed effects, which control for any residual influence of display order despite randomization. We estimate this using ordinary least squares (OLS) regression and cluster standard errors at the participant level to account for repeated evaluations by the same individual.

Next, we examine how model type and perceived quality jointly influence consumer choice:

\begin{equation*}
\Pr(\text{SL}_{ijkt}=1) = \Lambda\!\left(
\beta_1 \text{UA}_k + \beta_2 \text{Rate}_{ijkt} + \gamma' P_k
\right),
\end{equation*}

\noindent where $\Lambda(z)=\exp(z)/(1+\exp(z))$ maps the linear index into a choice probability, and $\text{SL}_{ijkt}$ equals one if participant $i$ selects image $k$ in choice set $j$, and zero otherwise. The coefficient $\beta_1$ captures the direct effect of the Utility-Aware Generator on choice probability, holding perceived quality constant, while $\beta_2$ captures the relationship between perceived image quality and the likelihood of selection. Including both terms allows us to assess whether the effect of the Utility-Aware Generator operates primarily through improved perceived quality or also reflects additional factors not captured by the rating measures. As in the rating model, we include position fixed effects and cluster standard errors at the participant level. In both specifications, $i$ indexes participants, $j$ indexes choice sets, and $k$ indexes images.

Table~\ref{tab:main_results} reports the estimation results. For the rating outcomes, columns (1)--(2) estimate OLS models with position fixed effects and standard errors clustered at the participant level. Across both measures, images generated by the Utility-Aware Generator receive significantly higher evaluations. In column (1), the Utility-Aware Generator increases perceived realism and trustworthiness by 0.708 points on a 7-point scale ($p < .001$). Similarly, in column (2), it improves perceived professional quality by 0.741 points ($p < .001$). These effects are economically meaningful, representing approximately a 16--17\% increase relative to the baseline mean.

Columns (3)--(4) examine purchase choice. In column (3), the Utility-Aware Generator significantly increases the likelihood that an image is selected for purchase. In column (4), the coefficient remains positive and significant after controlling for perceived realism and professional quality, suggesting that the Utility-Aware Generator improves choice not only by increasing perceived quality, but also through additional image attributes valued by consumers.

\begin{table}[t]
\centering
\caption{Effect of Using Utility-Aware Generator on Image Quality and Choice}
\label{tab:main_results}
\begin{tabular}{lcccc}
\toprule
D.V. & (1) Realism & (2) Professional & (3) Purchase & (4) Purchase \\
\midrule
Utility-Aware Generator & 0.708*** & 0.741*** & 1.038*** & 0.786*** \\
 & (0.081) & (0.078) & (0.133) & (0.137) \\

Realism &  &  &  & 0.739*** \\
 &  &  &  & (0.110) \\

Professional &  &  &  & 0.229** \\
 &  &  &  & (0.094) \\

Position FE & Y & Y & Y & Y \\
\midrule
Obs. & 2832 & 2832 & 2832 & 2832 \\
\bottomrule
\end{tabular}\\
\vspace{0.5em}
\footnotesize{
Notes: Q2 = realism \& trustworthiness; Q3 = professional quality. 
Columns (1)--(2) use OLS; columns (3)--(4) use logit. 
Standard errors clustered at the participant level in parentheses. 
$^* p<.1$, $^{**} p<.05$, $^{***} p<.01$.
}
\end{table}
\subsection{Study II: Validating Utility-Aware Editing for Airbnb Property Images}

We next examine whether utility-aware editing improves consumer evaluations in a hospitality setting. Participants viewed an original Airbnb room image together with AI-edited alternatives generated by Stable Diffusion, OpenAI, Flux, and the Utility-Aware Generator. For each edited image, participants rated realism and trustworthiness (Q1), uniqueness (Q2), aesthetic appeal (Q3), and willingness to book (Q4) on 7-point Likert scales. We recruited 117 participants from Prolific.com. Three participants were removed because of
insufficient completion time. After excluding observations with missing responses, the final image-level sample consists of 347 evaluations.

Table~\ref{tab:airbnb_means} reports model-free evidence. The Utility-Aware Generator achieves the highest average ratings across all four outcomes, followed by Flux and OpenAI, while Stable Diffusion performs substantially worse. The largest gains appear in realism and booking likelihood, suggesting that utility-aware editing improves both perceived credibility and commercial appeal.

We estimate the following image-level specification separately for each outcome:
\begin{align*}
Q_{isk}^{(m)} = \beta_m \text{UA}_k + \gamma_s + \epsilon_{isk},
\end{align*}
where $m\in{1,2,3,4}$ denotes question, $i$ indexes participants, $s$ indexes image sets, and $k$ indexes images. $\text{UA}_k$ is an indicator for the Utility-Aware Generator, and $\gamma_s$ denotes set fixed effects. Standard errors are clustered at the participant level. 

Table~\ref{tab:airbnb_reg} reports the results. The Utility-Aware Generator significantly improves all four outcomes: realism by 1.13 points ($p<.001$), uniqueness by 0.50 points ($p<.01$), aesthetic appeal by 0.72 points ($p<.001$), and willingness to book by 0.77 points ($p<.001$). These magnitudes are economically meaningful on a 7-point scale, indicating that utility-aware editing improves both perceived image quality and downstream consumer preference.

\begin{table}[t]
\centering
\caption{Model-Free Average Ratings by Model: Airbnb Study}
\label{tab:airbnb_means}
\begin{tabular}{lcccc}
\toprule
Model & Realism & Uniqueness & Aesthetic & Booking \\
\midrule
Flux & 5.145 & 4.627 & 4.964 & 4.807 \\
OpenAI & 4.198 & 4.527 & 4.989 & 4.582 \\
Stable Diffusion & 3.268 & 3.381 & 3.660 & 3.423 \\
Utility-Aware Generator & 5.263 & 4.671 & 5.263 & 5.039 \\
\bottomrule
\end{tabular}\\
\vspace{0.5em}
\footnotesize{
Notes: Mean ratings on a 7-point scale. Higher values indicate more favorable evaluations.
}
\end{table}

\begin{table}[t]
\centering
\caption{Effect of Utility-Aware Editing on Airbnb Image Evaluations}
\label{tab:airbnb_reg}
\begin{tabular}{lcccc}
\toprule
 D.V. & (1) Realism & (2) Uniqueness & (3) Aesthetic & (4) Booking \\
\midrule
Utility-Aware Generator 
& 1.130*** & 0.503*** & 0.717*** & 0.767*** \\
& (0.183) & (0.169) & (0.159) & (0.169) \\
\midrule
Image Set FE & Yes & Yes & Yes & Yes \\
Observations & 347 & 347 & 347 & 347 \\
\bottomrule
\end{tabular}\\
\vspace{0.5em}
\footnotesize{
Notes: OLS estimates with participant-clustered standard errors in parentheses. 
The independent variable is an indicator for images generated by the Utility-Aware Generator (baseline = other models). 
Set fixed effects are included. 
$^* p<.1$, $^{**} p<.05$, $^{***} p<.01$.
}
\end{table}

\section{Discussion and Conclusion}

This section summarizes the main findings, contributions, and managerial implications of the paper. We first revisit the research questions and summarize the core empirical and behavioral results. We then discuss the methodological, theoretical, and empirical contributions of the utility-aware framework. Finally, we highlight how sellers, hosts, and platforms can use utility-aware image generation to improve marketplace content while avoiding over-stylized or unrealistic visual outputs.

\subsection{Conclusion and Contributions}

Product images are economically valuable assets on digital platforms because they shape consumer attention, trust, and purchasing behavior \citep{zhang2022airbnb,li_xie_2020}. Although generative AI and text-to-image models have substantially improved firms' ability to produce visual content at scale, existing systems remain primarily optimized for image--text coherence rather than marketplace performance. As a result, generated images may be visually aligned with prompts but still fail to improve consumer demand or economic outcomes.

This paper addresses this gap by proposing a modular and flexible utility-aware framework for multimodal image generation that incorporates estimated consumer demand into contrastive representation learning. We introduce a Utility-Aware InfoNCE objective that integrates demand-driven visual attributes into standard image--text alignment, yielding a Utility-Aware CLIP model that can be used to guide image generation toward economically meaningful outputs.

Our contributions are threefold.
\textit{First}, methodologically, we develop a general utility-aware contrastive learning framework that embeds estimated demand parameters into multimodal representation learning. Importantly, the framework is modular. It acts as a plug-in layer that can be embedded into existing and emerging text-to-image systems to improve their direct commercial usage. This enables utility-guided generation without redesigning the underlying generative AI architecture. \textit{Second}, theoretically, we establish a discrete-choice interpretation of contrastive learning and show how utility components enter the InfoNCE objective. This provides an economic foundation for CLIP-style models and clarifies how incorporating demand-based attributes preserves the information-maximization principle while shifting learned representations toward economically relevant visual features. \textit{Third}, empirically, we apply the framework to image generation in Amazon and Airbnb settings and evaluate performance using demand-based metrics and human-subject experiments. Across both contexts, the Utility-Aware Generator consistently outperforms state-of-the-art benchmarks, including Stable Diffusion, GPT-Image, and Flux, by producing images that better balance realism, semantic consistency, and marketplace effectiveness, with behavioral evidence from human experiments confirming improvements in consumer choice.

\subsection{Managerial Implications}

For sellers and hosts, the framework offers a scalable alternative to prompt engineering and manual design by using observed demand data to guide image generation. For instance, Amazon sellers can generate additional usage-context images while preserving product identity and avoiding overly stylized scenes that reduce realism. Similarly, Airbnb hosts can enhance room images by adjusting lighting and minor decorative elements while maintaining layout authenticity.

For platforms, the approach enables systematic listing improvement, content auditing, and counterfactual image testing. Platforms can recommend alternative visuals, identify underperforming listings, and support sellers with limited design resources by reducing reliance on professional photography or iterative prompting. Importantly, the results highlight that ``more aesthetic'' is not always better, as demand often follows an inverted U-shaped relationship with visual attributes. Overly polished or stylized images may reduce trust and perceived authenticity. Utility-aware generation helps balance visual appeal with realism and product fidelity.

The utility-aware framework also offers practical advantages over alternative adaptation approaches such as LoRA (Low Rank Adaptation) \citep{hu2022lora}  or full fine-tuning of the underlying generative model. Fine-tuning-based methods require curating a large set of high-quality images, retraining billions of parameters in the base diffusion model, and repeating this process whenever the underlying model is updated. In contrast, our framework operates as a lightweight guidance layer on top of the existing generative model, which makes it substantially faster and cheaper to deploy. It is also more inclusive in the set of attributes it can incorporate: rather than implicitly learning preferences from a fixed set of training images, the utility component can flexibly encode multiple demand-driven features simultaneously, including colorfulness, brightness, symmetry, aesthetic quality, and any additional attributes a firm wishes to add as long as they can be measured. This makes the framework easier to update as demand evidence evolves, easier to audit because the utility functions are explicit and interpretable, and easier to embed into different base models as the generative AI landscape continues to change.

\subsubsection*{Guidance for Choosing the Utility Weight.}
The weight $\eta$ in the Utility-Aware Contrastive Loss governs the trade-off between demand-related visual attributes and semantic alignment, with higher $\eta$ yielding higher-demand generation. We discuss two approaches:
\begin{itemize}
\item \textit{Experience-based calibration (no demand data).} Without historical demand data, $\eta$ and utility coefficients are set from managerial expertise: category managers encode attribute preferences (e.g., inverse-U structures for brightness) and tune $\eta$ via A/B testing or sample review. This suits new sellers, new categories, or platforms lacking systematic demand data.
\item \textit{Demand-based calibration (with demand data).} When sales outcomes link to visual content, $\eta$ is calibrated empirically by estimating demand regressions and selecting the value that maximizes predicted demand while maintaining acceptable fidelity on a validation sample. This suits established sellers with rich behavioral data.
\end{itemize}

Firms can combine both, beginning with experience-based priors and transitioning to demand-based calibration as data accumulate. The framework is modular, acting as a plug-in layer compatible with state-of-the-art models that consistently steers outputs toward commercial effectiveness.

\bibliographystyle{apalike} \bibliography{refs}
\end{document}